\documentclass{article}

\usepackage{spconf,amsmath,graphicx,hyperref}

\usepackage{booktabs} 
\usepackage{multirow}
\usepackage{amssymb}

\newif\ifarxivversion

\arxivversiontrue

\title{SCALING FORCED ALIGNMENT TO END-USER DEVICES}
\name{Lawry Sorenson \qquad Michael Crandall \qquad Eric K Ringger \qquad Stephen D Richardson}
\address{Brigham Young University \\ Computer Science Department \\ Provo, UT, United States}
\begin{document}
%
\maketitle

\begin{abstract}
The Viterbi algorithm has been previously used to perform forced alignment of audio to text to mine training data from online resources. However, many existing implementations have quadratic time and space complexity, scaling poorly to long input sequences. We propose two optimizations to address this issue. First, we apply the Hirschberg algorithm to perform the alignment in place using linear memory. Second, we model the alignment between speech and text as a constrained random walk, allowing us to prune the search space with arbitrary confidence while accounting for transcription errors. The Hirschberg optimization reduces memory usage from 140 GB to 5 MB for three-hour inputs while producing identical alignments in one-third the time of torchaudio when both run on a CPU. We achieve an additional 2x speedup with pruning on inputs longer than 20 minutes while preserving alignment accuracy in more than 98\% of tested cases.
\end{abstract}
\begin{keywords}
speech recognition, forced alignment, Hirschberg--Viterbi algorithm
\end{keywords}
\section{Introduction}
\label{sec:intro}

Despite recent advances, model performance for low-resource languages remains underdeveloped. Their advancements notwithstanding, Large Language Models (LLMs) still suffer from this issue, especially for languages with poor tokenizer support. This issue limits access to information and cross-lingual communication for minority-language communities. 

This problem is particularly prevalent in speech processing, as there is little speech data available in many languages. A common method to address data scarcity has used forced alignment (FA) on public data sources, such as New Testament readings or parliamentary recordings. However, recent FA implementations have $\mathcal{O}(n^2)$ time and memory complexity, which limits their application to large inputs. Our work focuses on improving the scalability of FA so that it can run on long input sequences with hardware available to most end users. We make our implementation available as an open-source Python package. \footnote{\url{https://github.com/byu-matrix-lab/hirschberg-viterbi}}

\section{Related Work}
\label{sec:relatedwork}

Early work to align multilingual speech with text relied heavily on rule-based transfer between languages, frequently requiring pronunciation dictionaries of individual words. The CMU Wilderness dataset by Black \cite{black2019cmu} represented early efforts to scale speech technology to a wide variety of languages. Their methodology generalized the approach of Prahallad et al. \cite{prahallad2007automatic}, which proposed FA with dynamic programming to match audio with phonetic transcriptions of text.

Since then, Pratap et al. \cite{pratap2024scaling} trained speech recognition models for more than 1,000 languages. Their methodology used a shared romanization to map each script into the Latin alphabet. They trained a large, multilingual Automatic Speech Recognition (ASR) model (Massively Multilingual Speech, or MMS) on the romanized scripts, which they used to bootstrap rough alignments in new languages.

However, both approaches rely on multilingual transliteration tools, which Black \cite{black2019cmu} argues degrade alignment performance for low-resource languages, as pronunciation tends to be language-dependent. Kürzinger et al \cite{kurzinger2020ctc} introduced ctc-segmentation (ctc-seg), proposing FA with models trained with Connectionist Temporal Classification (CTC) loss without pronunciation dictionaries or transliteration. \cite{graves2006connectionist}

However, the FA implementation in ctc-seg often scaled poorly when applied to especially long input sequences, as considering every possible alignment requires $\mathcal{O}(n^2)$ time and memory. Kürzinger et al. \cite{kurzinger2020ctc} proposed limiting alignments to a window around the diagonal to drop the complexity to $\mathcal{O}(nw)$, where $w$ is the window size. The Kaldi speech recognition toolkit \cite{povey2011kaldi} with beam-search pruning has also been adapted to FA, \cite{mcauliffe2017montreal} but existing work has not explored what the pruning parameters should be, and newer alignment implementations fall back to the $\mathcal{O}(n^2)$ approach. \cite{pratap2024scaling,rastorgueva2023nemo}

The $\mathcal{O}(n^2)$ algorithm has significant memory requirements, and aligning an hour-long audio segment with its transcription can easily exceed 16 GB of RAM. Pratap et al. \cite{pratap2024scaling} managed to work around the issue of running the algorithm on a GPU with insufficient memory by pushing backtracking data out to the CPU RAM. The FA implementation for MMS has been included in torchaudio \cite{yang2022torchaudio}, and some independent work has already been done to reduce its memory requirements. We note ctc-forced-aligner, which uses 2 bits to store backtracking data instead of a full byte and boasts a 5x memory reduction from torchaudio. \cite{ashraf2024ctc}


One approach that has been taken to address memory usage in the similar problem of aligning two strings to find their longest common subsequence is Hirschberg's algorithm. \cite{hirschberg1975linear} Hirschberg's algorithm uses a divide-and-conquer approach to align two strings effectively in-place. This approach has not been applied to FA, but we demonstrate that the same principles work here as well.

\section{Datasets}
\label{sec:datasets}

Our main development dataset was composed of General Conference interpretation data from The Church of Jesus Christ of Latter-day Saints, which we refer to as CJCLDS-GC. This dataset consists of live interpretation of discourses assisted by pretranslations into 73 languages supported by MMS, most of which have 100+ hours of audio. The average duration of each discourse audio is 11.7 minutes.\ifarxivversion We describe the dataset more in Appendix \ref{sec:datasetfacts}.\fi

Although all the data is available for browsing online,\footnote{\label{footnote:gc-dataset}\url{https://churchofjesuschrist.org/study/general-conference}} this dataset is not currently released for easy research use. For reproducibility, we also evaluate our forced-alignment optimizations on two other corpora. The first is the Buckeye corpus, which consists of 40 hour-long interviews in English that have been segmented into 255 10-minute audio files and labeled at a phone level. \cite{pitt2007buckeye} The second is EuroSpeech, which consists of sentence-aligned transcription-audio pairs from parliamentary recordings of 22 European countries. \cite{pfisterer2025eurospeech} EuroSpeech has recordings from more than 20k parliament sessions, with an average duration of 2.6 hours each.

\section{Methodology}
\label{sec:method}

We developed two optimizations for the Viterbi algorithm using the CJCLDS-GC dataset before evaluating them both on the Buckeye and EuroSpeech datasets.

\subsection{Hirschberg--Viterbi Algorithm}
\label{ssec:hirschberg}

First, we implemented Hirschberg's algorithm for FA to support $\mathcal{O}(n^2)$ runtime with $\mathcal{O}(n)$ memory. This algorithm applies the principles of divide-and-conquer and meet-in-the-middle and only requires that the alignment can be computed forward and backward through the data. The algorithm works from both sides to compute the optimal alignment at the midpoint in the audio before recursing. Scores are tracked in-place to limit memory usage. We end recursion once the subproblems become sufficiently small to fit within 1 KB, which we found limits overhead from excessive recursion.

\subsection{Search Space Pruning}
\label{ssec:pruning}

\begin{figure}[t]
\centering
\includegraphics[width=0.9\linewidth]{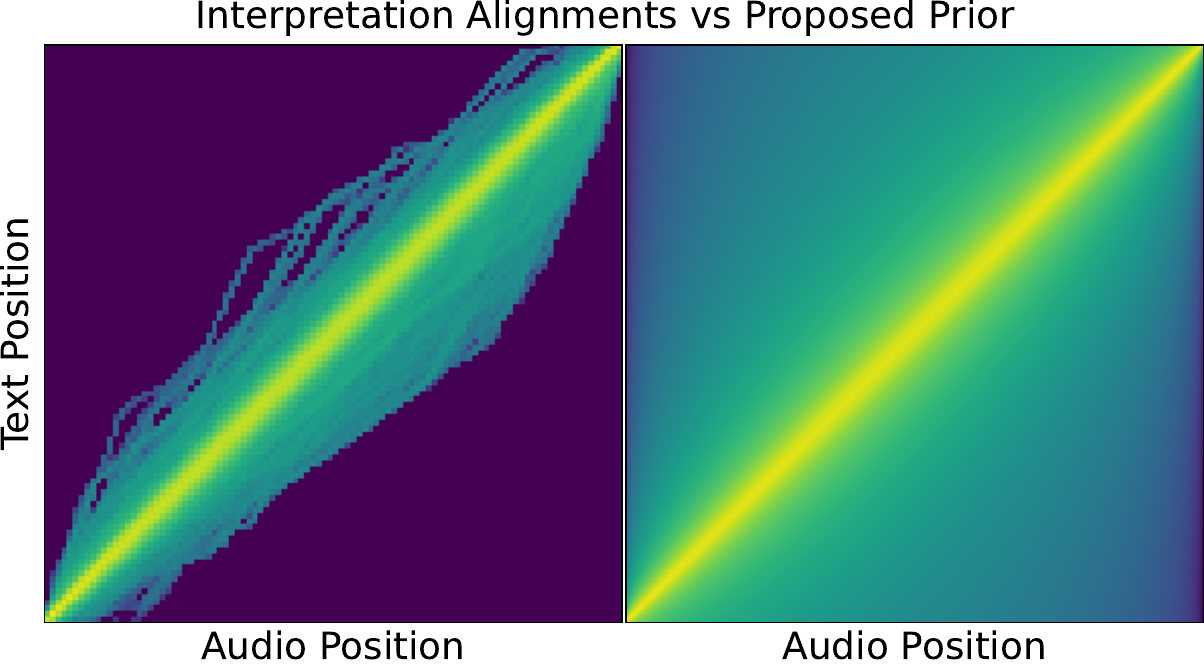}
\caption{Logarithmic heatmaps showing a random sample of alignments from the CJCLDS-GC dataset (left), and the alignment prior that we use to predict FA pruning bounds (right). This prior is created from the assumption that phoneme durations are randomly distributed and stationary.}
\label{img:prop-prior}
\end{figure}

In the second optimization, we limit the search space to a window around the diagonal, and we propose a theoretical basis for its size. While Kürzinger et al. \cite{kurzinger2020ctc} already proposed using a fixed-width window during FA to reduce the space and runtime complexity from $\mathcal{O}(n^2)$ to $\mathcal{O}(n)$, we argue that the width of such a window should scale with the input size.

We compute an upper bound for the desired window size by adding a prior indicating that alignments tend to be along the diagonal. To create this prior, we model the alignment as a random walk of phoneme durations bounded on both ends by the start and end of the sequence. To account for variations in speaking rate across speakers and audio segments, we scale the phoneme durations for each audio such that the expected length of the transcription matches the duration of the audio.

From this prior, we form a normal approximation to the alignment position of a single character based on the central limit theorem, which we present as Equation \ref{eq:pos_prior}\ifarxivversion and derive in Appendix \ref{sec:prior-deriv}\fi. Here $\mu_l$ and $\sigma_l$ are the expected value and standard deviation of the duration of the audio for the text to the left of the current character, while $\sigma_r$ is the standard deviation of the text to the right. We illustrate this prior in Figure \ref{img:prop-prior}, a logarithmic heatmap showing the probability that each character occurs at each timestep under this model. For reference, the same figure also shows a random selection of 50 text-audio alignments from the CJCLDS-GC dataset. We assume the variance of each character is within 3.7 times its mean, based on an analysis of the CJCLDS-GC dataset.

\begin{equation}\label{eq:pos_prior}
p_c \sim \mathcal{N}(\mu_l, (\sigma^{-2}_l + \sigma^{-2}_r)^{-1})
\end{equation}

We treat identifying the window size as finding a confidence interval for the alignment of each character, yet this only represents the confidence of a single point being in the window and does not accurately represent the confidence that all points along the alignment are in the window. By the union bound, we apply a conservative correction by adjusting the confidence for each window size to $1 - \delta / n$, where $\delta$ is the target error rate and $n$ is the transcription length. We set the minimum window size to 15 seconds, as the model breaks down at the edges of the audio.

In our data exploration, we note that most of the deviations from the diagonal are due to inaccurate transcriptions or periods of silence. These outliers include \href{https://www.churchofjesuschrist.org/study/general-conference/2022/10/58nelson?lang=eng}{sharing videos during a discourse}, \href{https://www.churchofjesuschrist.org/study/general-conference/2020/10/46nelson?lang=ell}{the interpreter finishing reading the prepared translation two minutes before the original speaker}, and \href{https://www.churchofjesuschrist.org/study/general-conference/1999/04/your-light-in-the-wilderness?lang=eng}{the speaker inviting others to speak with them}. As such, we modify the pruning bounds to account for both by estimating the accuracy of the transcription in terms of precision and recall. This is similar to the character error rate, but we treat insertions and deletions separately, as they affect the expected alignment position in opposite directions. Our pruning model can also fold silence into the recall parameter, as it is part of the audio that does not correspond to the text.

Our system evaluates the worst-case alignment given lower bounds for the transcription precision and recall. That is, we compute the lower bound for a character's expected position in time by assuming that all false insertions are before the current character and all false deletions are after it, and we do the opposite to compute the upper bound. Using a lower bound for these parameters accounts for not knowing the transcription accuracy during inference. 

We look at the width of the predicted window to evaluate the time complexity. The variance of $p_c$ is proportional to $n$, and the standard deviation is proportional to $\sqrt{n}$. The adjusted confidence then scales based on the decay rate of the inverse complementary error function, $\text{erfc}^{-1}(1/n)$, which is sub-logarithmic. When precision and recall are similar, the adjustment can shift the position of a character within the audio by $2\epsilon$, where $\epsilon$ is the error rate. This effectively increases the search space by $\mathcal{O}(n^2\epsilon)$, giving a loose time complexity

\begin{equation}
\mathcal{O}(n \sqrt n \log n + n^2 \epsilon)
\end{equation}

\begingroup
\setlength{\tabcolsep}{1.5pt} 
\begin{table*}[ht]
\centering

\label{tab:pruningperf}
\begin{tabular}{|c|l|cc|cc|cc|cccc}
\toprule
\multicolumn{2}{|c|}{} & \multicolumn{2}{c|}{\textbf{Buckeye}}
& \multicolumn{2}{c|}{\textbf{CJCLDS-GC}}
& \multicolumn{2}{c|}{\textbf{EuroSpeech}}
& \multicolumn{4}{c}{\textbf{Overall}} \\
\multicolumn{2}{|c|}{\textbf{Method}}
& Exact & $<250$ ms
& Exact & $<250$ ms
& Exact & $<250$ ms
& Failure & Time & Max Time & Max Mem. \\
\midrule

\multicolumn{2}{|c|}{Unpruned HV} & 100\% & 100\% & 100\% & 100\% & 100\% & 100\% & 0\% & 59:36:39 & 2:50:44 & 22.4 MB \\

\midrule
\multirow{3}{*}{\rotatebox{90}{Uncut}}
& HV + VAD Hint & \textbf{100\%}$\dagger \ddagger$ & \textbf{100\%}$\dagger \ddagger$ & \textbf{98.2\%}$\dagger \ddagger$ & 98.3\% $\ddagger$ & \textbf{100\%} $\dagger \ddagger$ & \textbf{100\%} $\ddagger$ & 0\% & 32:38:36 & 1:22:26 & 41.5 MB \\
& Kaldi & 67.5\%$\ddagger$ & 75.0\% $\ddagger$ & 93.8\%$\ddagger$ & \textbf{99.5\%}$\ast \ddagger$ & 69.6\% $\ddagger$ & 94.2\%$\ddagger$ & 0.03\% & 287:58:12 & $>$3 days & 21.5 GB \\
& ctc-seg & 0\% & 52.5\% & 0\% & 70.0\% & 0\% & 18.8\% & 6.0\% & 10:15:45 & 1:04:20 & $>$80 GB \\

\midrule
\multirow{3}{*}{\rotatebox{90}{VAD}}
& HV & 0\% & \textbf{77.5\%}$\ddagger$ & \textbf{1.97\%}$\dagger\ddagger$ & 93.4\%$\ddagger$ & 0\% & 82.0\%$\ddagger$ & 0\% & 11:40:39 & 27:23 & 1.9 GB \\
& Kaldi & 0\% & 70.0\% & 1.93\%$\ddagger$ & \textbf{94.6\%}$\ast \ddagger$ & 0\% & \textbf{82.6\%}$\ddagger$ & 0.03\% & 254:46:38 & $>$3 days & 17.0 GB \\
& ctc-seg & 0\% & 57.5\% & 0\% & 69.5\% & 0\% & 70.3\% & 5.0\% & 9:16:38 & 59:00 & $>$80 GB \\

\bottomrule

\end{tabular}
\caption{Pruning performance for different methods with their default parameters. We include metrics when VAD is used to cut periods of silence before processing. Runtime and memory usage is aggregated across all datasets. The longest input across the test sets is 16 hours. Systems marked with $\ast$, $\dagger$, and $\ddagger$ significantly outperformed our system, Kaldi, and ctc-seg, respectively, with p-values under $0.05$.}
\end{table*}
\endgroup

\section{Evaluation}

\subsection{Pruning Accuracy}

While traditional evaluation for FA tends to look at the distance between the predicted timestamps and the ground truth, our optimizations will ideally result in the same predicted alignment as the base Viterbi algorithm. We propose a stricter metric by measuring the percentage of input audio files for which the pruned alignment perfectly matches the unpruned alignment. To also allow for slight errors, we also report the percentage of files where pruning shifted the timestamps by less than 250 ms on average.

We compare our pruning method with Kaldi and ctc-seg on the three datasets. For a fair comparison, we use the default settings for each method. The pruning in ctc-seg starts by limiting the alignment space to a window of 8000 timesteps centered on the diagonal and doubles this window size whenever backtracking throws an index-out-of-bounds error. Kaldi uses beam-search, pruning beams at each timestep that have a log probability of 16 lower than the current best. We add an outer loop that doubles this bound whenever the Kaldi alignment fails. The default parameters for our pruning are 97\% transcription accuracy and 99\% alignment confidence.

We also explore the accuracy of pruning when each method is paired with pyannote Voice Activity Detection (VAD) \cite{bredin23pyannote2} to remove silence in the audio before processing. When not removing the silence, we pass a hint of the estimated amount of silence into our pruning calculation. We add VAD to the runtime of each method that uses it.

We limit each alignment to 80 GB of RAM and 3 days of processing time. Alignments that exceed those bounds are marked as failures. We only use the EuroSpeech and CJCLDS-GC test sets (10\% of each dataset), but we use all 40 interviews in the Buckeye dataset. We use MMS to produce the logits for FA and present our results in Table 1.

Since VAD introduces the potential for cascaded errors, we also evaluate our pruning bounds using the human-labeled periods of silence in the Buckeye dataset. We found that once periods of silence are removed from the audio, our pruning bounds perfectly contained the dataset alignment while assuming a perfect transcription at 99\% pruning confidence. \footnote{We found 3 audio files with mislabeled periods of silence in the dataset, which we corrected for our tests. These periods occurred from time 522-528 near the end of file 2903b, from 78-104 in file s0503b, and from 206-222, 497-503, and 504-518 in file s4003b.}

\begin{figure*}
\centering
\includegraphics[width=0.8\linewidth]{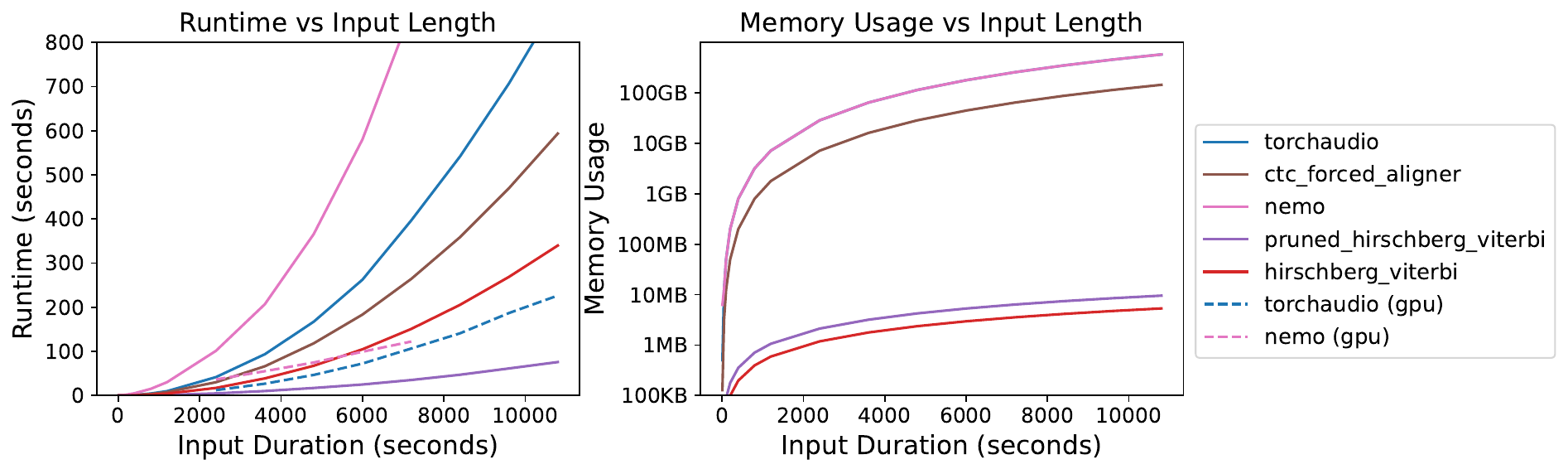}
\caption{Runtime and memory usage of various implementations of the Viterbi algorithm. Note that torchaudio and NeMo have the same memory usage and overlap in the graph. The GPU used has only 80 GB of VRAM, so that line for NeMo ends early once it no longer has enough VRAM to run.}
\label{img:emp-complex}
\end{figure*}

\subsection{Complexity Analysis}

Lastly, we evaluate the runtime and memory efficiency of our FA optimizations against other implementations that do not use pruning. We compare our code to the existing implementations in torchaudio, NeMo, and ctc-forced-aligner when running on both a CPU and GPU. Our own implementation is a CPU-based PyTorch extension. We do not include Kaldi and CTC-Seg in this section because we compared against them in the last section, and their runtime and memory usage can vary widely for two inputs of the same length, while our pruning has consistent scaling for set pruning options.

We evaluate each system on randomly generated inputs of various durations, where logits have 50 frames per second and the transcriptions have 12.7 characters per second---the average in the CJCLDS-GC dataset. We run each implementation 10 times per input duration. We track memory using the max RSS for the running process. We average the performance across runs and plot them in Figure \ref{img:emp-complex}. \footnote{Tracking memory usage with RSS is imperfect as it does not account for swap memory. When memory usage was too small to detect a change in RSS, we estimate it using the allocations in the code.} CPU computation was performed on a single core of an AMD EPYC 7763 (2.45 GHz), while GPU computation was done on an NVIDIA A100.

\section{Discussion}

Though Kaldi slightly outperformed our pruning method at producing close approximations to the baseline in a few cases, even our unpruned alignment was dramatically faster than Kaldi and used less memory. Our code was slower than ctc-segmentation, but had far less memory usage and better pruning accuracy.

We looked through the instances where Kaldi outperformed our system and found that they generally occurred when the transcription was less than 97\% accurate. In all the cases we looked at, this was due to \href{https://www.churchofjesuschrist.org/study/general-conference/2017/10/the-book-of-mormon-what-would-your-life-be-like-without-it?lang=eng}{the published text having additional content that was not in the audio}. We expect that our pruning would handle these if the accuracy bound is adequately decreased to account for the mismatch. We argue that our pruning parameters are far more interpretable to end users than those of Kaldi or ctc-seg. To use our code, users need only give a lower bound for the transcription accuracy, but for ctc-seg they must give an upper bound of the window width in timesteps, or for Kaldi, an upper bound on the log probability deviation between the optimal and greedy decodings.

Our implementation uses a contiguous copy of its input, which is what causes the large jump in memory usage when slicing the audio based on VAD. VAD also introduces cascaded errors (like other chunk-based alignment systems) \cite{pfisterer2025eurospeech}, while ctc-seg uses a different recurrence than CTC loss, which is why those exact match rates are so low. 

Interestingly, we found that the Hirschberg optimization significantly reduced runtime despite requiring some recomputation. We expect that the significant reduction in memory usage allowed most of the algorithm to run within the cache.

\ifarxivversion
In comparison to existing algorithms, the NeMo Forced Aligner is the only implementation we assessed that currently supports batch computation of alignments, and we suggest that it be used when aligning many short segments on a GPU. Torchaudio on a GPU remains the fastest unpruned implementation when sufficient RAM is available. On a CPU, our implementation was the fastest and used the least memory for all inputs. Our Hirschberg--Viterbi implementation could align three hours of audio on one CPU core in 271.3 seconds with 5.2 MB of RAM, while torchaudio running on a GPU needed 220.4 seconds and 144 GB of RAM. When both implementations ran on a CPU, the Hirschberg--Viterbi implementation consistently ran in one-third of the time of torchaudio. Our pruned Hirschberg--Viterbi implementation can do the same three-hour alignment in 57.2 seconds on a CPU for high-accuracy transcriptions.
\fi

\section{Conclusion}

We propose a theoretical basis for computing window size, novelly led by interpretable transcription accuracy, which, combined with the existing Hirschberg algorithm, significantly improves the scalability of FA. We hope that this work contributes to research efforts for low-resource language communities, especially by allowing communities to process their own data without significant hardware requirements.

\ifarxivversion

\section{Limitations and Future Work}

We note that the pruning method we propose is blind to the data, while Kaldi's pruning is entirely based on the data. We expect that the two could be combined where Kaldi's method prunes based on the data that has already been processed, and ours prunes based on expectations for the data that has not been processed yet.

Our model currently assumes that the speaking rate is consistent, which we find is not always the case. Modeling variable speaking rate would require wider pruning bounds and could likely be done by constraining the expected amount of variation in speaking rate. The bounds could then be created from the worst cases in which the speaking rate monotonically increases or decreases during the entire audio file.

Lastly, we note that current methods of FA identify a single alignment between audio and text. We note that the Hirschberg--Viterbi algorithm we use could be slightly modified to efficiently calculate the location of key points in the text (such as sentence boundaries) while considering all possible paths between them. We suspect this approach could improve alignment accuracy similar to beam search in decoding, but we also leave it to future work.

\else
\vfill\pagebreak
\fi

\section{Compliance with Ethical Standards}

This study was performed retrospectively on publicly available data. We received explicit permission to use the CJCLDS-GC for our own research, but we do not release the dataset publicly at this time due to concerns from the Church about interpreter privacy. Nevertheless, the data may be inspected online, as specified in footnote \ref{footnote:gc-dataset}.

\section{Funding Acknowledgments}

No funding was received for conducting this study. The authors have no relevant financial or nonfinancial interests to disclose.

\bibliographystyle{IEEEbib}
\bibliography{strings,refs}

\ifarxivversion

\appendix

\section{Prior Derivation}
\label{sec:prior-deriv}

We start by analyzing the dataset to compute the mean and variance in grapheme durations per language. As CTC loss is peaky, we measure the distance between one peak and the next as the duration of the grapheme. \cite{huang2024peaky} Although graphemes and phonemes frequently do not have a one-to-one matching, we assume that there is strong correlation between graphemes and audio duration.

To account for changes in speaking speed, we scale all audio times such that the means are 1 so that we can compute the relative standard deviation in grapheme duration. We do this as we expect the standard deviation to be linearly affected by changes in speaking rate. We found that in many languages with phonemic scripts, the average standard deviation was 1.6 times the mean duration, while in languages with logographic characters (Chinese, Japanese Kanji) the average standard deviation was 0.95 times the mean. We expect specific graphemes to have much smaller standard deviations than this, but we computed these statistics for all characters together to avoid statistical overfitting. We use a standard deviation ratio of 1.9 in our experiments as the upper bound across the languages in our development set.

By the linearity of expectations, the expected duration of a sequence of graphemes will be the sum of the individual means. If we assume that individual graphemes are independently distributed, then the distribution of the duration of such a sequence will approach a normal distribution whose variance is the sum of individual variances by the Central Limit Theorem (CLT). Importantly, this assumption is not necessarily true, as changes in speaking rate will affect adjacent graphemes. However, CLT can also hold when the sequence is strongly mixing, i.e. adjacent events are dependent, but the dependence between distant events decays to zero. \cite{billingsley1995probability} We expect speech to be strongly mixing, and we also assume that speaking rates remain relatively constant, primarily to maintain that the distributions are stationary. We do find outliers with variable speaking rates, but our model holds for most of the evaluated datasets.

We continue with the assumption that the distribution of the duration of a spoken sequence of graphemes approaches the normal distribution. For any character in the sequence, we can now model its position in the audio by merging the distributions of the durations on both sides. Given a point in the grapheme sequence, its expected location in the audio is along the diagonal. The probability of a deviation from said diagonal is the joint probability that the durations of the sequences on both sides of that point deviate the same amount from their means. Here we again assume that the distributions of the two sides are independent, which we note has the same issues stated above. We start with the distributions of the durations of the left and right sequences:

\begin{equation}
d_l \sim \mathcal{N}(\mu_l, \sigma^2_l), d_r \sim \mathcal{N}(\mu_r, \sigma^2_r)
\end{equation}

Note that we scaled the prior such that $\mu_l + \mu_r$ is the duration of the audio sequence. The mean location for the character position $c$ is then at time $\mu_l$. From there, the distribution then depends on the probability that two normal distributions have the same deviation. We solve for the resulting distribution by temporarily shifting the mean to 0 and taking the product of two normal distributions. For simplicity, we express the normalization constants as $\beta_l$ and $\beta_r$.

\begin{equation}
\mathbb{P}(x) = \beta_l \exp(-\frac{x^2}{2\sigma^2_l}) \beta_r \exp(-\frac{x^2}{2\sigma^2_r})
\end{equation}

Note that we can add exponents and factor out the $-\frac{x^2}{2}$ terms.

\begin{equation}
\mathbb{P}(x) = \beta_l \beta_r \exp(-\frac{x^2}{2}(\frac{1}{\sigma^2_l}+\frac{1}{\sigma^2_r}))
\end{equation}

Note that this is conveniently similar to the normal distribution. Renormalizing this distribution results in the following distribution for the position of character $c$, represented as $p_c$.

\begin{equation}
p_c \sim \mathcal{N}(\mu_l, (\sigma^{-2}_l + \sigma^{-2}_r)^{-1})
\end{equation}

This distribution represents the probability that the alignment passes through a specific time at character $c$. Critically, this model breaks down towards the edges of the transcription and audio as there are not enough characters on both sides for the central limit theorem to apply. We partially address this issue by setting the window size to the maximum of the predicted size or 15 seconds. We have not researched other sizes for the edge windows.

\section{Other Visualizations}

Figure \ref{img:hirschberg-vis} shows the recursive nature of the Hirschberg algorithm that allows it to prune half of the search space at each level of recursion

\begin{figure}
\centering
\includegraphics[width=\linewidth]{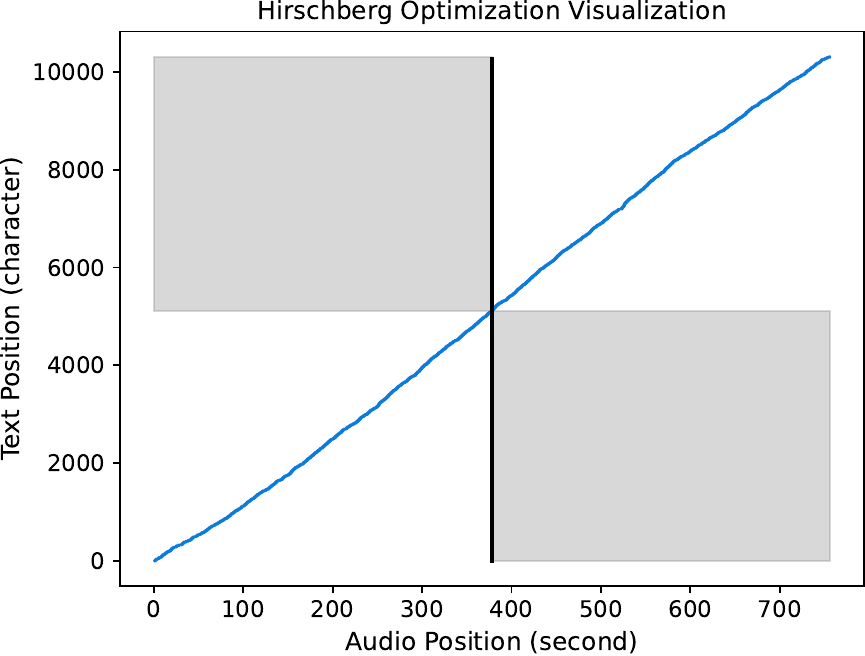}
\caption{
Visualization of the Hirschberg algorithm applied to forced alignment. The algorithm works from both sides to compute an optimal pivot point in the middle, then recurses to the lower left and upper right areas in the figure.
}
\label{img:hirschberg-vis}
\end{figure}

Figure \ref{img:pruning-error} depicts one of the corrections for transcription inaccuracy in predicted pruning bounds. The alignment source for this figure is available \href{https://www.churchofjesuschrist.org/study/general-conference/2023/10/51nelson?lang=bul}{here}.

\begin{figure}
\centering
\includegraphics[width=0.9\linewidth]{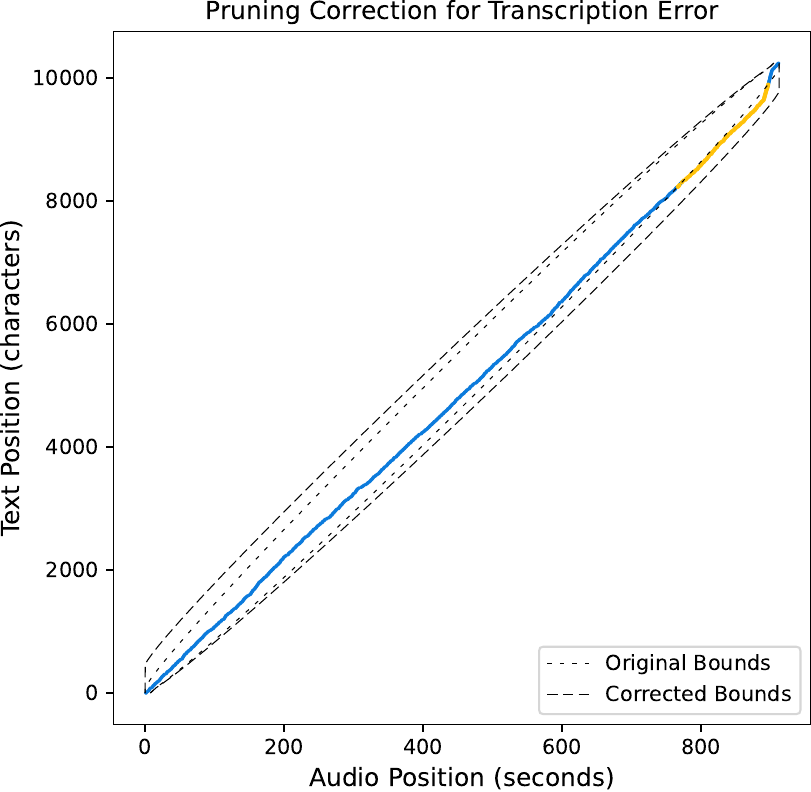}
\caption{
In this specific alignment, which is for the Bulgarian interpretation of Russell Nelson's discourse "Think Celestial!" from 2023, the interpreter did not read a list of cities at the end of the discourse, instead using the original English audio. The city names are in the transcription, but there is no Bulgarian audio corresponding to them. The list of city names occupies 3.7\% of the transcription, and adjusting the pruning bound for 96.3\% transcription precision produces the outer pruning bounds, which fully contain the alignment.
}
\label{img:pruning-error}
\end{figure}

\begin{figure}
\centering
\includegraphics[width=\linewidth]{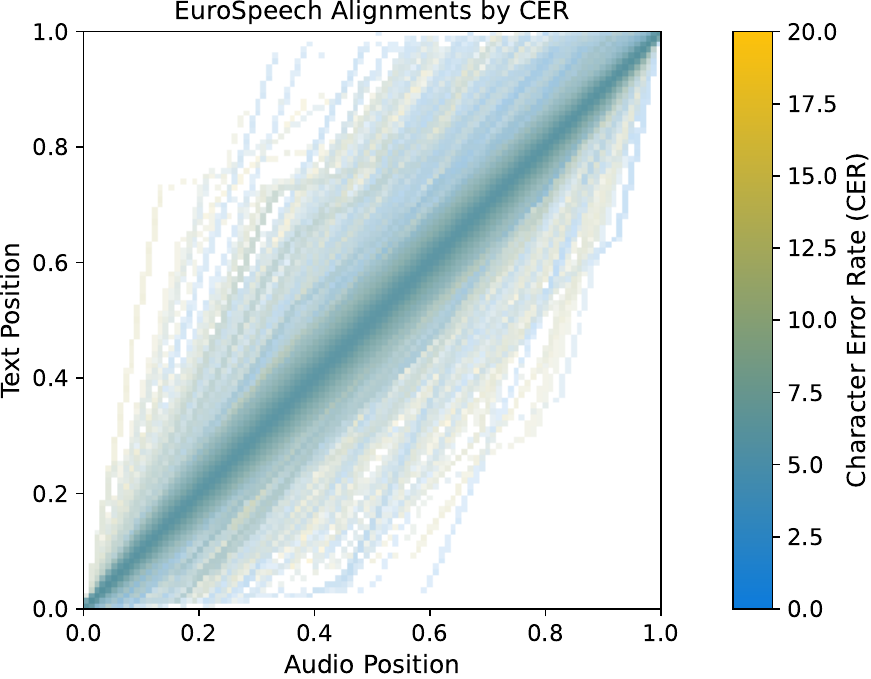}
\caption{The alignments in the EuroSpeech dataset after removing audio between labeled sentences. We color the heatmap using the CER of the transcription. Hue depicts CER, while color intensity indicates logarithmic frequency in the dataset. We note that alignments further from their expected position tend to have higher CER.}
\label{img:eurospeech-align}
\end{figure}

\begin{figure*}[ht]
\centering
\includegraphics[width=0.8\linewidth]{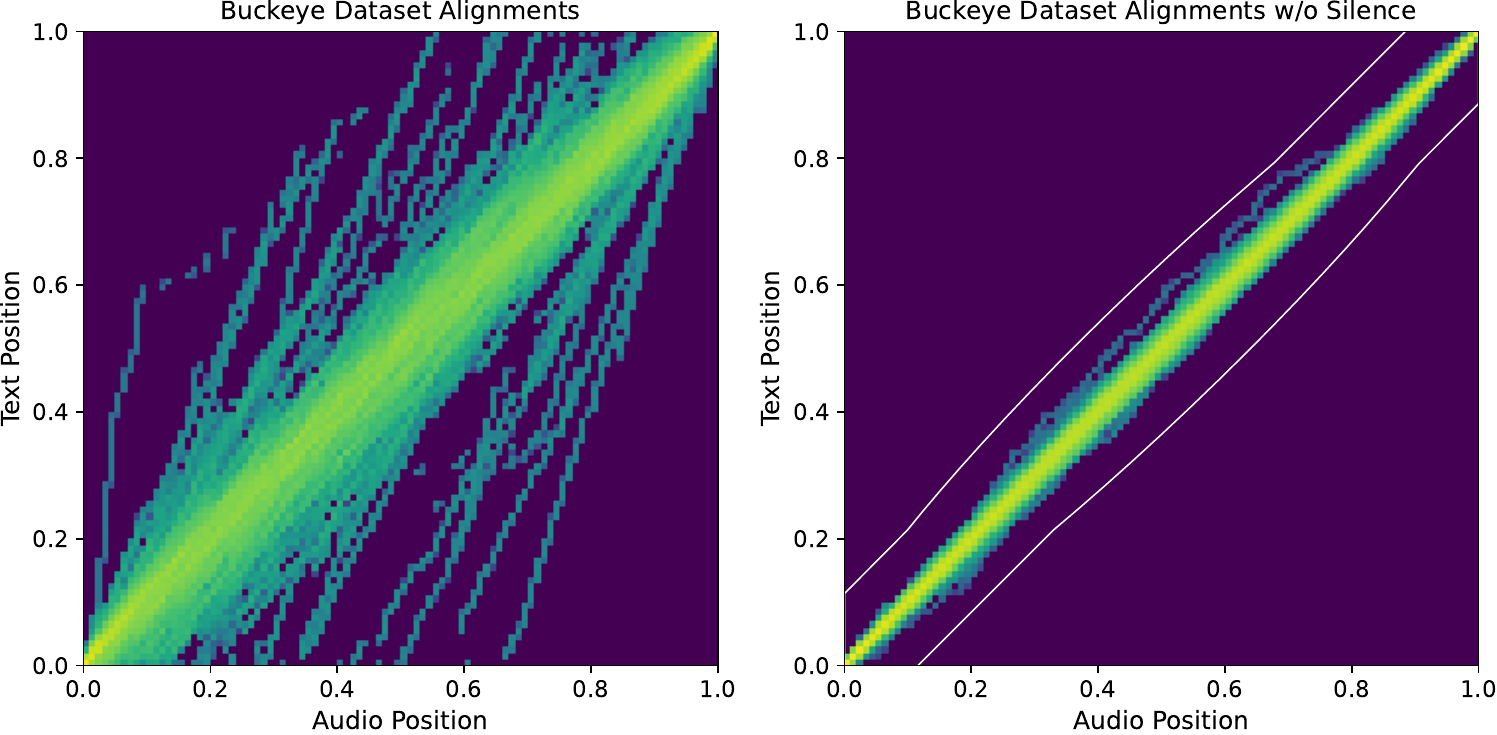}
\caption{After removing labeled periods of silence from the Buckeye dataset, all transcription alignments were contained within their predicted pruning space. The lines on the right depict the pruning space for s2002a, one of the shorter recordings with a wider relative pruning space. Note that these depict the segmented interview labels available with dataset, not our reconstruction of the 40 interviews.}
\label{img:buckeye-results}
\end{figure*}

The alignments in the Buckeye dataset are visualized in Figure \ref{img:buckeye-results}, where we depict the change in alignment after 
removing periods of labeled silence.

\begin{table*}[h]
\centering
\label{tab:datasetstats}
\begin{tabular}{|l|rrrrr|}
\toprule
Dataset & Languages & Files & Total Duration (h) & Minimum Length & Maximum Length \\
\midrule

Buckeye & 1 & 40 & 38 & 39:25 & 1:17:54 \\
CGCLDS-GC & 85 & 63961 & 12509 & 00:04 & 41:28 \\
--- Our Test Split & 73 & 5766 & 1128 & 01:03 & 32:16 \\
EuroSpeech & 22 & 27683 & 77031 & 00:10 & 17:42:45 \\
--- Test Split & 21 & 140 & 794 & 00:22 & 16:20:36 \\

\bottomrule

\end{tabular}
\caption{The size of each of the datasets used in our study. Note that this details our reconstructions of the unsegmented audio files for each dataset. We use the full Buckeye dataset in our tests, both only evaluate our pruning on the test sets for CGCLDS-GC and EuroSpeech. The CGCLDS-GC dataset contains the data through 2023, when we took a snapshot of the data at the time. There is one fewer language in the EuroSpeech test set, becuase EuroSpeech on HuggingFace is missing a test set for Italian. There are 12 fewer languages in our test set from the CGCLDS-GC dataset, because we limit our experiments to languages supported by MMS. }
\end{table*}

We also visualize the sentence-level alignments in the full EuroSpeech dataset given the timestamps in the metadata for each setence. We use this data to reconstruct the alignment of the full audio, again capping periods of silence between sentences at 250 ms. We note that the metadata does not label silence or pauses within a sentence, causing them to remain in our reconstruction. We graph these aligments against the CER for each transcript in Figure \ref{img:eurospeech-align}.

\section{Dataset Statistics}
\label{sec:datasetfacts}

Here we present Table 2, which details the number of languages and characteristics of the duration of audios for each of the datasets that we used in our experiments.

\fi

\end{document}